\documentclass[sigconf]{acmart}

\AtBeginDocument{%
  \providecommand\BibTeX{{%
    \normalfont B\kern-0.5em{\scshape i\kern-0.25em b}\kern-0.8em\TeX}}}

\setcopyright{acmcopyright}
\copyrightyear{2020}
\acmYear{2020}
\acmDOI{10.1145/1122445.1122456}

\acmConference[Petra '21]{Petra '21: The Pervasive Technologies Related to Assistive Environments}{June 29-- July 02, 2021}{Corfu, Greece}
\acmBooktitle{Petra '21: The Pervasive Technologies Related to Assistive Environments,
  June 29-- July 02, 2021, Corfu, Greece}
\acmPrice{15.00}
\acmISBN{978-1-4503-XXXX-X/18/06}

\begin{document}

\title{Automated system to measure Tandem Gait to assess executive functions in children}


\author{Mohammad Zaki Zadeh}
\affiliation{%
  \institution{University of Texas at Arlington}
  \city{Arlington}
  \country{USA}}
\email{mohammad.zakizadehgharie@mavs.uta.edu}

\author{Ashwin Ramesh Babu}
\affiliation{%
  \institution{University of Texas at Arlington}
  \city{Arlington}
  \country{USA}}

\author{Ashish Jaiswal}
\affiliation{%
  \institution{University of Texas at Arlington}
  \city{Arlington}
  \country{USA}}

\author{Maria Kyrarini}
\affiliation{%
  \institution{University of Texas at Arlington}
  \city{Arlington}
  \country{USA}}

\author{Morris Bell}
\affiliation{
  \institution{Yale University}
  \city{New Haven}
  \country{USA}}

\author{Fillia Makedon}
\affiliation{
  \institution{University of Texas at Arlington}
  \city{Arlington}
  \country{USA}}


\renewcommand{\shortauthors}{Zaki Zadeh, et al.}

\begin{abstract}
As mobile technologies have become ubiquitous in recent years, computer-based cognitive tests have become more popular and efficient. In this work, we focus on assessing motor function in children by analyzing their gait movements. Although there has been a lot of research on designing automated assessment systems for gait analysis, most of these efforts use obtrusive wearable sensors for measuring body movements. We have devised a computer vision-based assessment system that only requires a camera which makes it easier to employ in school or home environments. A dataset has been created with 27 children performing the test. Furthermore in order to improve the accuracy of the system, a deep learning based model was pre-trained on NTU-RGB+D 120 dataset and then it was fine-tuned on our gait dataset. The results highlight the efficacy of proposed work for automating the assessment of children's performances by achieving $76.61\%$ classification accuracy.
\end{abstract}



\maketitle

\section{Introduction}

\begin{figure}[h]
  \centering
  \includegraphics[width=0.9\linewidth]{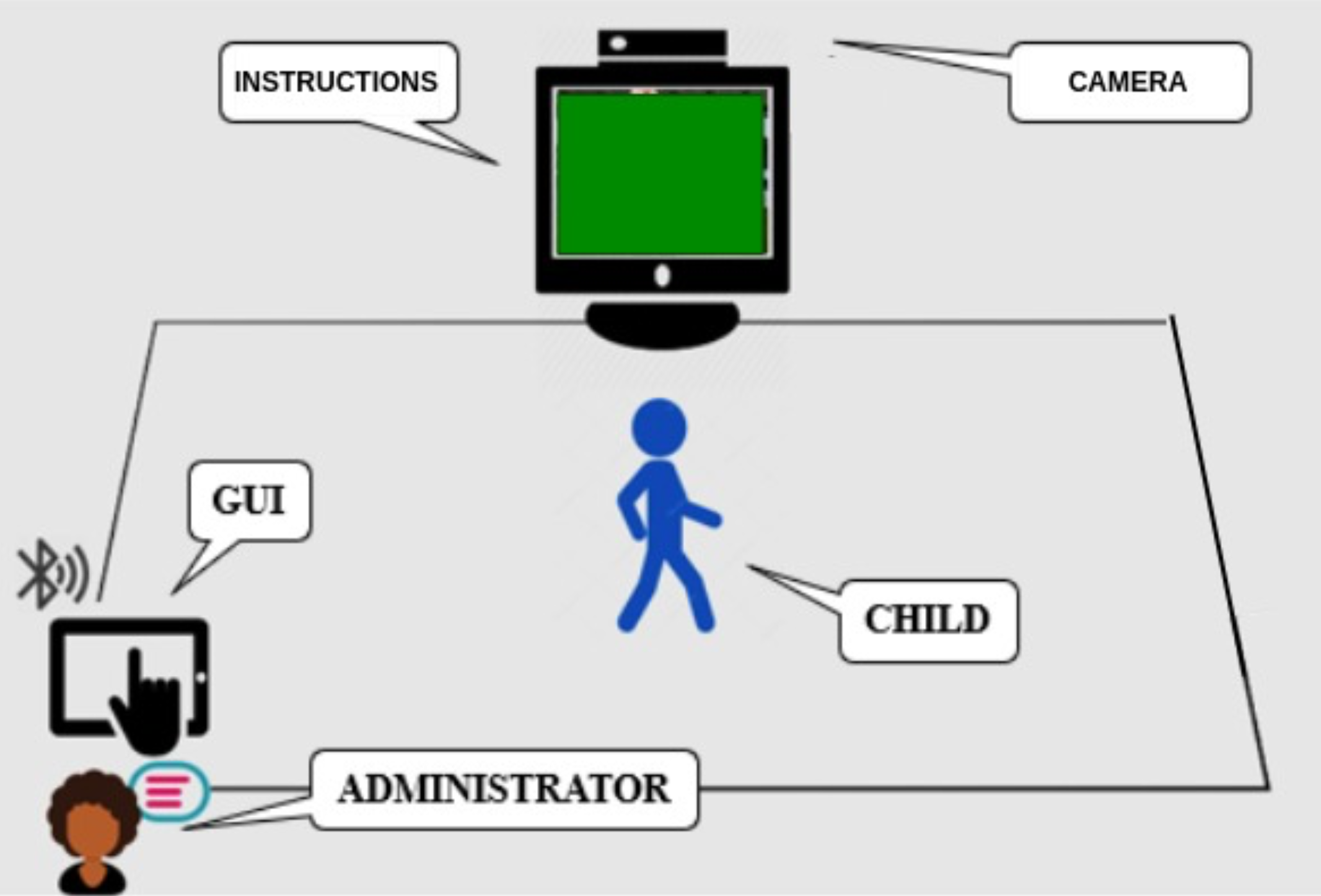}
  \caption{Data collection setup}
  \label{fig:atec}
\end{figure}

\begin{figure*}[h]
  \centering
  \includegraphics[width=1.0\linewidth]{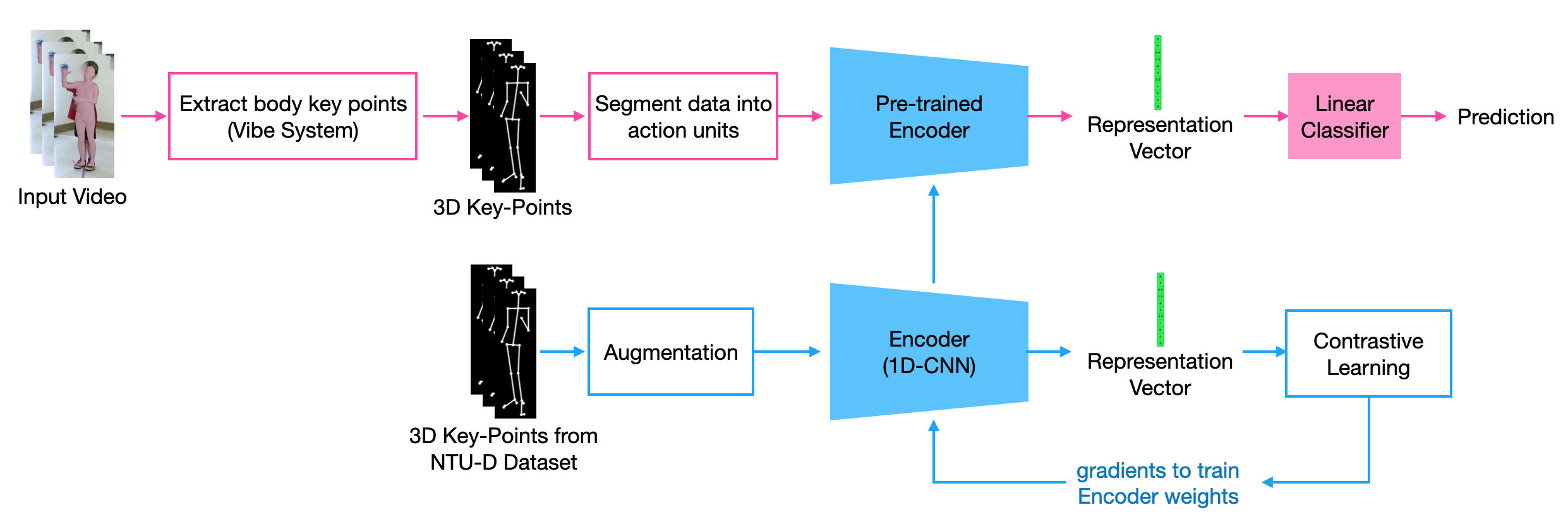}
  \caption{Proposed method architecture. Top (pink): Supervised classification. Bottom (blue): self-supervised pre-training}
  \label{fig:arch}
\end{figure*}

Executive functions are high-order mental processes that enable us to successfully plan, multitask, focus, remember instructions, switch tasks, coordinate etc. forming the foundation of the cognitive development.  They mostly rely on brain functions such as working memory, mental flexibility, motor skills, and others.  Motor skills is one of the important skills that humans learn during their childhood.  They generally involve movements of the large muscles in the arms, legs, and torso.  Humans rely on these skills for their every day activities at home, work/school, and in the community.  Children affected with neurological conditions such as ADHD  exhibit motor abnormalities \cite{leitner2007gait, buderath2009postural}, especially when it comes to balance.  Such impairments when not treated at the right time can affect everyday activities of a person that in turn  can affect other functions such as fine motor skills.  Building an automated system to access such disorders paves way for an efficient diagnosis and treatment.

The NIH toolbox is a standardized set of tests for cognitive assessment that empowers automated assessment through sensors and mobile applications.  Specifically, there are numerous tests to assess balance which require sensors such as accelerometer that is attached to the body to measure it.  The overall goal of this work is to build a low-cost automated assessment system that uses computer vision to analyse participants performing the task and score them on the basis of standard cognitive measures such as gait and balance.  

In this work (Figure \ref{fig:arch}), the focus of the automated assessment system is on the "Tandem gait" task which is part of a larger system called ATEC-Activated Test of Embodied Cognition \cite{iwoar2019, case19, icmi20}. A dataset has been created with 27 children performing the gait task. In order to automatically evaluate subject's performance, first VIBE \cite{vibe} human pose estimation system was used to extract 3D body key-points. Then a deep learning based model was trained to classify subject's steps as valid or invalid. Furthermore in order to improve the accuracy of the system, the model was pre-trained on NTU-RGB+D 120 dataset and then fine-tuned on our gait dataset. Contrastive learning \cite{mdpi20} framework was employed to pretrian the model in self-supervised manner. The results shows that pre-trained model can achieve satisfying results even when small amount of annotated data is available for training.

The rest of the paper is structured as follows: Section 2 discusses the related work, Section 3 explains the setup for data collection with information about the gait dataset. Section 4 describes the proposed method along with the results. Finally, conclusion and future works are mentioned in section 5.

\begin{figure}[ht]
  \centering
  \includegraphics[width=1.0\linewidth]{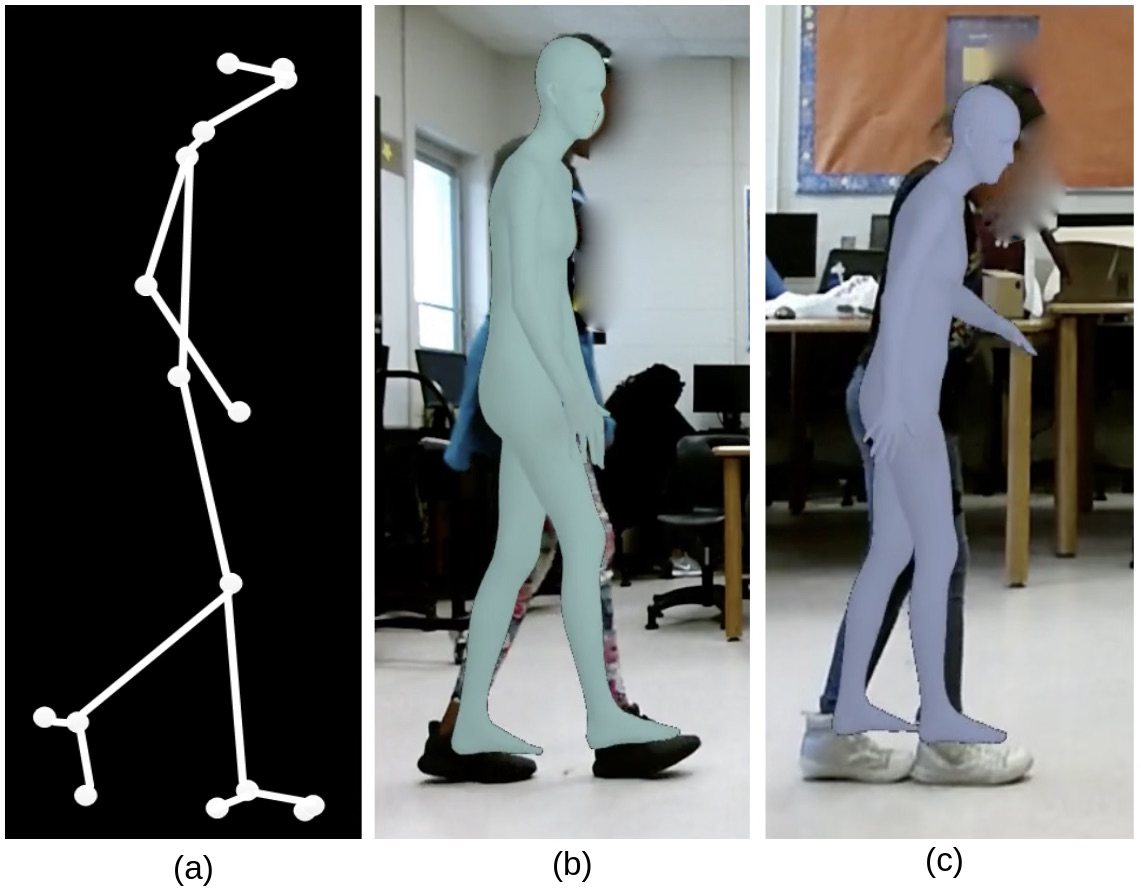}
  \caption{(a): Skeleton key points, (b) Example of a invalid step, (c): Example of an valid step}
  \label{fig:gait}
\end{figure}

\section{Related Works}

There has been a plethora of research in recent years that tackle the problem of analysing body gait for prediction and diagnosis of multiple disorders. In \cite{gait16}, machine learning methods have been widely used for gait assessment through the estimation of spatio-temporal parameters. The proposed methodology was tested on gait data recorded on two pathological populations (Huntington’s disease and post-stroke subjects) and healthy elderly controls. They used data from inertial measurement units placed at shank and waist. In \cite{gait17}, wearable sensor technologies were employed for development of new methods for monitoring parameters that characterize mobility impairment such as gait speed outside the clinic. In their work, authors try to extend these methods that are often validated using normal gait patterns to subjects with gait impairments.

In \cite{gait18}, the focus was on diagnosis of Vascular Dementia during or prior to vascular cognitive impairment. They explored using gait analysis which include stride length, lateral balance, or effort exerted for a particular class of activity. Although gait has clear links to motor activities, they investigate an interesting link to visual processing since the visual system is strongly correlated with balance. Various gait metrics have been investigated, and their potential to identify vascular cognitive impairment has been evaluated. In \cite{gait19}, the issue of support for diabetic neuropathy (DN) recognition is addressed. In this research, gait biomarkers of subjects is used to identify people suffering from DN. To achieve this, a home-made body sensor network was employed to capture raw data of the walking pattern of individuals with and without DN. The information was then processed using three sampling criteria and 23 assembled classifiers in combination with a deep learning algorithm.

In \cite{gait19b}, the effects of human fatigue due to repetitive and physically challenging jobs that cause Work-related Musculoskeletal Disorder (WMSD) was investigated. This study was designed to monitor fatigue through the development of a methodology that objectively classifies an individual’s level of fatigue in the workplace by utilizing the motion sensors embedded in smartphones. Using Borg’s Ratings of Perceived Exertion (RPE) to label gait data, a machine learning algorithms was developed to classify each individual’s gait into different levels of fatigue. Finally, in \cite{gait20}, the aim of the study was to determine whether gait and balance variables obtained with wearable sensors could be utilized to differentiate between Parkinson’s disease and essential tremor.

\begin{figure}[h]
  \centering
  \includegraphics[width=0.8\linewidth]{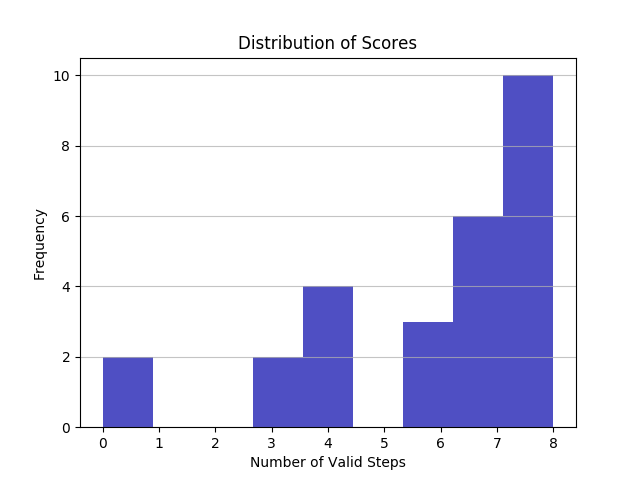}
  \caption{Distribution of children scores (number of valid steps)}
  \label{fig:hist}
\end{figure}

\begin{table}
\begin{center}
\begin{tabular}{|l|c|c|c|}
\hline
Method & 80\% & 50\% & 10\% \\
\hline\hline
Supervised & $72.39$ & $63.33$ & $52.13$ \\
Contrastive Learning (E2E)& $76.61$ & $72.44$ & $70.90$\\
Contrastive Learning (MoCo) & $76.61$ & $74.03$ & $72.46$\\
\hline
\end{tabular}
\end{center}
\caption{Gait Task: Top 1 classification accuracy of different methods for different train/test split. 80\% corresponds to using 80\% of datastet for traing and remaining for testing.}
\label{results}
\end{table}

\section{Dataset Description}

In this section, the data collection setup along with the characteristics of dataset are explained. Figure \ref{fig:atec} represents our video-based data collection setup. An RGB camera was used to collect the side view of the child performing the task. The recording modules were connected to an android-based interface which was controlled by the administrator (parents, teachers) to monitor the flow of the assessment.

Data was collected from children between the age of 6-10 across multiple school in the United States. Participants were invited to perform the assessment task in a classroom environment after parents consenting and completing the screening procedure required by the study protocol. A total of 27 recordings from 27 children were collected. In each recording, the kid was asked to perform 8 valid steps. A step is considered valid only if the heel of one foot touches the toe of another foot. Then subject's 3D body key-points were extracted using VIBE system \cite{vibe}. VIBE (Video Inference for Body Pose and Shape Estimation) is a video pose and shape estimation method that predicts the parameters of SMPL body model for each frame of an input video. From these key-points 17 of them including head, hands, hip, feet and toes were selected. 
Finally, the extracted data were divided into 8 equal segments (with overlap), each corresponding to an step. An example of a valid and an invalid step are presented in Figure \ref{fig:gait}. In these figures, children's body are covered by their estimated SMPL body mesh in order to see VIBE system body pose estimation in action and also to protect their privacy. 

In order to evaluate the performance of our system, each task performed by a child was manually scored by our assistants. These scores later acted as the ground truth for our algorithm. 
In Figure \ref{fig:hist}, the distribution of scores from different children is depicted. Here, the score is equal to the number of valid steps performed by a child. 

In order to pre-train the classifier model, publicly available NTU-RGB+D 120 \cite{ntu1,ntu2} were used. This dataset contains 120 action classes and 114,480 video samples. In this work only 3D skeletal data were employed. Similar to gait dataset, 17 equivalent key-points (head, hands, hip, feet and toes) were selected.

\begin{figure}[ht]
  \centering
  \includegraphics[width=\linewidth]{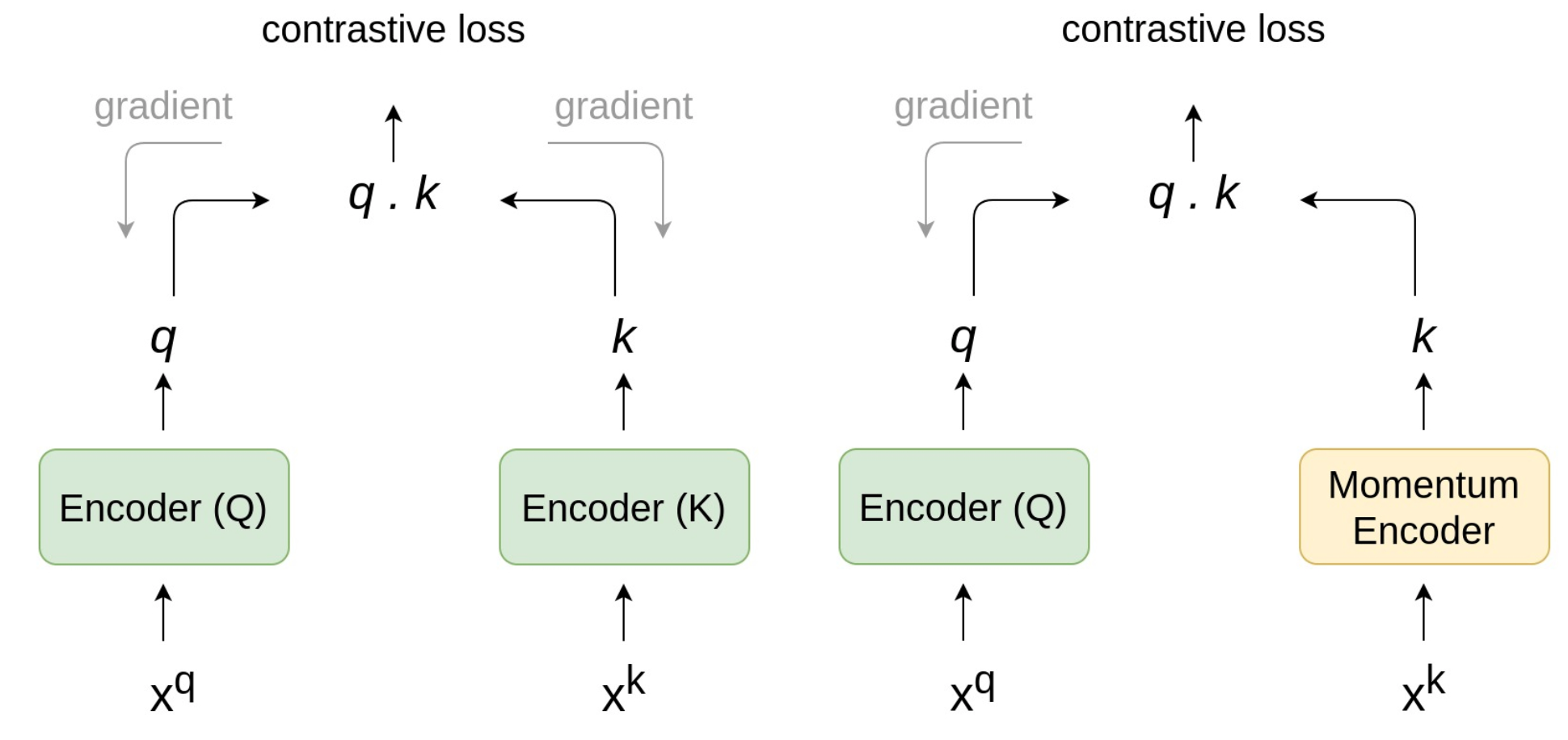}
  \caption{Different contrastive learning architecture ($q$ stand for query and $k$ for key) . Left: End-to-End training of encoders (E2E). Right: Using a momentum encoder as a dynamic dictionary lookup (MoCo). \cite{mdpi20}}
  \label{fig:cl}
\end{figure}

\section{Methods and Results}

After 3D body joints were extracted form input videos, they were divided into 8 segment with equal size. Each segment ($X \in {\mathbb{R}}^{32\times51}$) includes 32 samples with 51 features. The feature are x,y,z coordinates for each 17 key-points rasterized into one vector. Then, input was fed into an encoder network to obtain the compact representation $z \in {\mathbb{R}}^{256}$. Finally a linear classifier is used to classify input segment unto valid and invalid segments (Figure \ref{fig:arch}). In this work, a 4 layer 1D Convolutional Neural Network (CNN) \cite{alex} is used as encoder network.

For evaluating the performance of proposed methods in case of small amount of annotated data three scenarios were defined. In first scenario, $80\%$ of data was used for training and remaining $20\%$ for testing. In second scenario, $50\%$ of data was used for training and other $50\%$ for testing. Finally for scenario 3, $10\%$ of data was used for training and remaining $90\%$ for testing. The average classification accuracy was calculated by cross-validation. The results for baseline supervised method is shown in first row of Table \ref{results}. It is clear from the results that the baseline method classification accuracy decreases as training set becomes smaller. 

In order to improve the performance of the proposed system, we tried to pre-train encoder network on large NTU-RGB+D 120 dataset by using self-supervised training \cite{iui20,mdpi20}. One of the most popular self-supervised approaches is contrastive learning (CL) \cite{mdpi20,moco}. CL tries to group similar samples closer and diverse samples far from each other. To achieve this, a similarity metric (cosine similarity) is used to measure how close two representations are from each other. Representations are obtained by feeding input data into an encoder network. During training, one sample (query $x^q$) from the training dataset is taken and a transformed version (or another view in NTU dataset) of the sample is considered as a positive sample (positive key $x^{k_+}$), and the rest of the samples are considered as negative samples (positive key $x^{k_-}$). Training encourages encoder network to differentiate positive samples from the negative ones. 

Since number of negative samples affect the performance of CL methods \cite{mdpi20}, different strategies are used for selecting a large number of negative samples. In this work, two of this strategies called E2E and MoCo are used and their architecture are depicted in Figure \ref{fig:cl}. In End-to-End learning (E2E) a large batch size is used and all the samples in the batch except for the query and one positive sample are considered as negative.
Because large batch sizes inversely affects the optimization
during training, one possible solution would be to maintain a separate dictionary known as memory bank containing representations of negative keys. However, since maintaining a memory bank during training is complicated, the memory bank
can be replaced by a Momentum Encoder. The momentum encoder (MoCo) \cite{moco} generates a dictionary as a queue of encoded keys with the current mini-batch enqueued and the oldest mini-batch dequeued. The momentum encoder shares the same parameters as the query encoder ($\theta_q$) and its parameters ($\theta_k$) gets updated based on the parameters of the query encoder. ($\theta_k = m\theta_k + (1-m)\theta_q, m\in[0,1)$: momentum coefficient)

All of the above methods were trained using Pytorch framework \cite{pytorch} for 100 epochs. Also ADAM \cite{adam} was employed as optimizer with learning rate: $1\times10^{-4}$, $\beta_1$: $0.5$ and $\beta_2$: $0.999$. Furthermore, for all contrastive learning methods temperature hyperparameter $\tau$ and momentum coefficient $\mu$ were chosen as $0.1$ and $0.999$ respectively.


\section{Conclusion and Future Works}
In this work, we presented a dataset that incorporates recordings from 27 children performing the Tandem Gait task. We also designed a computer vision system with acceptable precision that analyzes a child's performance by counting the number of valid steps performed in the task. Our proposed method performs well even in case of having access to small amount of annotated training data.
Applying the proposed approach on all different tasks defined in ATEC such as ball-drop \cite{iui20,icmi20}, finger-oppose \cite{case19}, etc., and finally designing a general framework for cognitive assessment of children will be focus of future works.

\begin{acks}
This  work  was  partially  supported  by  National  Science Foundation grants IIS 1565328 and IIP 1719031.
\end{acks}

\bibliographystyle{ACM-Reference-Format}
\bibliography{sample-base}


\begin{thebibliography}{20}


\ifx \showCODEN    \undefined \def \showCODEN     #1{\unskip}     \fi
\ifx \showDOI      \undefined \def \showDOI       #1{#1}\fi
\ifx \showISBNx    \undefined \def \showISBNx     #1{\unskip}     \fi
\ifx \showISBNxiii \undefined \def \showISBNxiii  #1{\unskip}     \fi
\ifx \showISSN     \undefined \def \showISSN      #1{\unskip}     \fi
\ifx \showLCCN     \undefined \def \showLCCN      #1{\unskip}     \fi
\ifx \shownote     \undefined \def \shownote      #1{#1}          \fi
\ifx \showarticletitle \undefined \def \showarticletitle #1{#1}   \fi
\ifx \showURL      \undefined \def \showURL       {\relax}        \fi
\providecommand\bibfield[2]{#2}
\providecommand\bibinfo[2]{#2}
\providecommand\natexlab[1]{#1}
\providecommand\showeprint[2][]{arXiv:#2}

\bibitem[\protect\citeauthoryear{Babu, Zakizadeh, Brady, Calderon, and
  Makedon}{Babu et~al\mbox{.}}{2019}]%
        {case19}
\bibfield{author}{\bibinfo{person}{Ashwin~Ramesh Babu},
  \bibinfo{person}{Mohammad Zakizadeh}, \bibinfo{person}{James~Robert Brady},
  \bibinfo{person}{Diane Calderon}, {and} \bibinfo{person}{Fillia Makedon}.}
  \bibinfo{year}{2019}\natexlab{}.
\newblock \showarticletitle{An Intelligent Action Recognition System to assess
  Cognitive Behavior for Executive Function Disorder}. In
  \bibinfo{booktitle}{\emph{2019 IEEE 15th International Conference on
  Automation Science and Engineering (CASE)}}. IEEE, \bibinfo{pages}{164--169}.
\newblock


\bibitem[\protect\citeauthoryear{Buderath, G{\"a}rtner, Frings, Christiansen,
  Schoch, Konczak, Gizewski, Hebebrand, and Timmann}{Buderath
  et~al\mbox{.}}{2009}]%
        {buderath2009postural}
\bibfield{author}{\bibinfo{person}{Paul Buderath}, \bibinfo{person}{Kristina
  G{\"a}rtner}, \bibinfo{person}{Markus Frings}, \bibinfo{person}{Hanna
  Christiansen}, \bibinfo{person}{Beate Schoch}, \bibinfo{person}{J{\"u}rgen
  Konczak}, \bibinfo{person}{Elke~R Gizewski}, \bibinfo{person}{Johannes
  Hebebrand}, {and} \bibinfo{person}{Dagmar Timmann}.}
  \bibinfo{year}{2009}\natexlab{}.
\newblock \showarticletitle{Postural and gait performance in children with
  attention deficit/hyperactivity disorder}.
\newblock \bibinfo{journal}{\emph{Gait \& posture}} \bibinfo{volume}{29},
  \bibinfo{number}{2} (\bibinfo{year}{2009}), \bibinfo{pages}{249--254}.
\newblock


\bibitem[\protect\citeauthoryear{Dillhoff, Tsiakas, Babu, Zakizadehghariehali,
  Buchanan, Bell, Athitsos, and Makedon}{Dillhoff et~al\mbox{.}}{2019}]%
        {iwoar2019}
\bibfield{author}{\bibinfo{person}{Alex Dillhoff},
  \bibinfo{person}{Konstantinos Tsiakas}, \bibinfo{person}{Ashwin~Ramesh Babu},
  \bibinfo{person}{Mohammad Zakizadehghariehali}, \bibinfo{person}{Benjamin
  Buchanan}, \bibinfo{person}{Morris Bell}, \bibinfo{person}{Vassilis
  Athitsos}, {and} \bibinfo{person}{Fillia Makedon}.}
  \bibinfo{year}{2019}\natexlab{}.
\newblock \showarticletitle{An automated assessment system for embodied
  cognition in children: from motion data to executive functioning}. In
  \bibinfo{booktitle}{\emph{Proceedings of the 6th international Workshop on
  Sensor-based Activity Recognition and Interaction}}. \bibinfo{pages}{1--6}.
\newblock


\bibitem[\protect\citeauthoryear{He, Fan, Wu, Xie, and Girshick}{He
  et~al\mbox{.}}{2019}]%
        {moco}
\bibfield{author}{\bibinfo{person}{Kaiming He}, \bibinfo{person}{Haoqi Fan},
  \bibinfo{person}{Yuxin Wu}, \bibinfo{person}{Saining Xie}, {and}
  \bibinfo{person}{Ross Girshick}.} \bibinfo{year}{2019}\natexlab{}.
\newblock \bibinfo{title}{Momentum Contrast for Unsupervised Visual
  Representation Learning}.
\newblock
\newblock
\showeprint[arxiv]{1911.05722}~[cs.CV]


\bibitem[\protect\citeauthoryear{Jaiswal, Babu, Zadeh, Banerjee, and
  Makedon}{Jaiswal et~al\mbox{.}}{2020}]%
        {mdpi20}
\bibfield{author}{\bibinfo{person}{Ashish Jaiswal},
  \bibinfo{person}{Ashwin~Ramesh Babu}, \bibinfo{person}{Mohammad~Zaki Zadeh},
  \bibinfo{person}{Debapriya Banerjee}, {and} \bibinfo{person}{Fillia
  Makedon}.} \bibinfo{year}{2020}\natexlab{}.
\newblock \bibinfo{title}{A Survey on Contrastive Self-supervised Learning}.
\newblock
\newblock
\showeprint[arxiv]{2011.00362}~[cs.CV]


\bibitem[\protect\citeauthoryear{Karvekar}{Karvekar}{2019}]%
        {gait19b}
\bibfield{author}{\bibinfo{person}{Swapnali Karvekar}.}
  \bibinfo{year}{2019}\natexlab{}.
\newblock \showarticletitle{Smartphone-based Human Fatigue Detection in an
  Industrial Environment Using Gait Analysis}.
\newblock


\bibitem[\protect\citeauthoryear{Khan, Madden, and Snyder}{Khan
  et~al\mbox{.}}{2018}]%
        {gait18}
\bibfield{author}{\bibinfo{person}{Arshia Khan}, \bibinfo{person}{Janna
  Madden}, {and} \bibinfo{person}{Kristine Snyder}.}
  \bibinfo{year}{2018}\natexlab{}.
\newblock \showarticletitle{Framework Utilizing Machine Learning to Facilitate
  Gait Analysis as an Indicator of Vascular Dementia}.
\newblock \bibinfo{journal}{\emph{International Journal of Advanced Computer
  Science and Applications}}  \bibinfo{volume}{9} (\bibinfo{date}{01}
  \bibinfo{year}{2018}).
\newblock
\urldef\tempurl%
\url{https://doi.org/10.14569/IJACSA.2018.090801}
\showDOI{\tempurl}


\bibitem[\protect\citeauthoryear{Kingma and Ba}{Kingma and Ba}{2014}]%
        {adam}
\bibfield{author}{\bibinfo{person}{Diederik~P. Kingma} {and}
  \bibinfo{person}{Jimmy Ba}.} \bibinfo{year}{2014}\natexlab{}.
\newblock \bibinfo{title}{Adam: A Method for Stochastic Optimization}.
\newblock
\newblock
\showeprint[arxiv]{1412.6980}~[cs.LG]


\bibitem[\protect\citeauthoryear{Kocabas, Athanasiou, and Black}{Kocabas
  et~al\mbox{.}}{2019}]%
        {vibe}
\bibfield{author}{\bibinfo{person}{Muhammed Kocabas}, \bibinfo{person}{Nikos
  Athanasiou}, {and} \bibinfo{person}{Michael~J. Black}.}
  \bibinfo{year}{2019}\natexlab{}.
\newblock \bibinfo{title}{VIBE: Video Inference for Human Body Pose and Shape
  Estimation}.
\newblock
\newblock
\showeprint[arxiv]{1912.05656}~[cs.CV]


\bibitem[\protect\citeauthoryear{Krizhevsky, Sutskever, and Hinton}{Krizhevsky
  et~al\mbox{.}}{2012}]%
        {alex}
\bibfield{author}{\bibinfo{person}{Alex Krizhevsky}, \bibinfo{person}{Ilya
  Sutskever}, {and} \bibinfo{person}{Geoffrey~E Hinton}.}
  \bibinfo{year}{2012}\natexlab{}.
\newblock \showarticletitle{ImageNet Classification with Deep Convolutional
  Neural Networks}.
\newblock In \bibinfo{booktitle}{\emph{Advances in Neural Information
  Processing Systems 25}}. \bibinfo{publisher}{Curran Associates, Inc.},
  \bibinfo{pages}{1097--1105}.
\newblock


\bibitem[\protect\citeauthoryear{Leitner, Barak, Giladi, Peretz, Eshel,
  Gruendlinger, and Hausdorff}{Leitner et~al\mbox{.}}{2007}]%
        {leitner2007gait}
\bibfield{author}{\bibinfo{person}{Yael Leitner}, \bibinfo{person}{Ran Barak},
  \bibinfo{person}{Nir Giladi}, \bibinfo{person}{Chava Peretz},
  \bibinfo{person}{Rena Eshel}, \bibinfo{person}{Leor Gruendlinger}, {and}
  \bibinfo{person}{Jeffrey~M Hausdorff}.} \bibinfo{year}{2007}\natexlab{}.
\newblock \showarticletitle{Gait in attention deficit hyperactivity disorder}.
\newblock \bibinfo{journal}{\emph{Journal of neurology}} \bibinfo{volume}{254},
  \bibinfo{number}{10} (\bibinfo{year}{2007}), \bibinfo{pages}{1330--1338}.
\newblock


\bibitem[\protect\citeauthoryear{Liu, Shahroudy, Perez, Wang, Duan, and
  Kot}{Liu et~al\mbox{.}}{2019}]%
        {ntu2}
\bibfield{author}{\bibinfo{person}{Jun Liu}, \bibinfo{person}{Amir Shahroudy},
  \bibinfo{person}{Mauricio Perez}, \bibinfo{person}{Gang Wang},
  \bibinfo{person}{Ling-Yu Duan}, {and} \bibinfo{person}{Alex~C. Kot}.}
  \bibinfo{year}{2019}\natexlab{}.
\newblock \showarticletitle{NTU RGB+D 120: A Large-Scale Benchmark for 3D Human
  Activity Understanding}.
\newblock \bibinfo{journal}{\emph{IEEE Transactions on Pattern Analysis and
  Machine Intelligence}} (\bibinfo{year}{2019}).
\newblock
\urldef\tempurl%
\url{https://doi.org/10.1109/TPAMI.2019.2916873}
\showDOI{\tempurl}


\bibitem[\protect\citeauthoryear{Mannini, Trojaniello, Cereatti, and
  Sabatini}{Mannini et~al\mbox{.}}{2016}]%
        {gait16}
\bibfield{author}{\bibinfo{person}{Andrea Mannini}, \bibinfo{person}{Diana
  Trojaniello}, \bibinfo{person}{Andrea Cereatti}, {and}
  \bibinfo{person}{Angelo Sabatini}.} \bibinfo{year}{2016}\natexlab{}.
\newblock \showarticletitle{A Machine Learning Framework for Gait
  Classification Using Inertial Sensors: Application to Elderly, Post-Stroke
  and Huntington's Disease Patients}.
\newblock \bibinfo{journal}{\emph{Sensors}}  \bibinfo{volume}{16}
  (\bibinfo{date}{01} \bibinfo{year}{2016}), \bibinfo{pages}{1--14}.
\newblock
\urldef\tempurl%
\url{https://doi.org/10.3390/s16010134}
\showDOI{\tempurl}


\bibitem[\protect\citeauthoryear{McGinnis, Mahadevan, Moon, Seagers, Sheth,
  Wright, Dicristofaro, Silva, Jortberg, Ceruolo, Pindado, Sosnoff, Ghaffari,
  and Patel}{McGinnis et~al\mbox{.}}{2017}]%
        {gait17}
\bibfield{author}{\bibinfo{person}{Ryan McGinnis}, \bibinfo{person}{Nikhil
  Mahadevan}, \bibinfo{person}{Yaejin Moon}, \bibinfo{person}{Kirsten Seagers},
  \bibinfo{person}{Nirav Sheth}, \bibinfo{person}{John Wright},
  \bibinfo{person}{Steve Dicristofaro}, \bibinfo{person}{Ikaro Silva},
  \bibinfo{person}{Elise Jortberg}, \bibinfo{person}{Melissa Ceruolo},
  \bibinfo{person}{Jesus Pindado}, \bibinfo{person}{Jacob Sosnoff},
  \bibinfo{person}{Roozbeh Ghaffari}, {and} \bibinfo{person}{Shyamal Patel}.}
  \bibinfo{year}{2017}\natexlab{}.
\newblock \showarticletitle{A machine learning approach for gait speed
  estimation using skin-mounted wearable sensors: From healthy controls to
  individuals with multiple sclerosis}.
\newblock \bibinfo{journal}{\emph{PLOS ONE}}  \bibinfo{volume}{12}
  (\bibinfo{date}{06} \bibinfo{year}{2017}), \bibinfo{pages}{e0178366}.
\newblock
\urldef\tempurl%
\url{https://doi.org/10.1371/journal.pone.0178366}
\showDOI{\tempurl}


\bibitem[\protect\citeauthoryear{Moon, Song, Sharma, Lyons, Pahwa, Akinwuntan,
  and Devos}{Moon et~al\mbox{.}}{2020}]%
        {gait20}
\bibfield{author}{\bibinfo{person}{Sanghee Moon}, \bibinfo{person}{Hyun-Je
  Song}, \bibinfo{person}{Vibhash Sharma}, \bibinfo{person}{Kelly Lyons},
  \bibinfo{person}{Rajesh Pahwa}, \bibinfo{person}{Abiodun Akinwuntan}, {and}
  \bibinfo{person}{Hannes Devos}.} \bibinfo{year}{2020}\natexlab{}.
\newblock \bibinfo{title}{Classification of Parkinson's disease and essential
  tremor based on gait and balance characteristics from wearable motion
  sensors: A data-driven approach}.
\newblock
\newblock
\urldef\tempurl%
\url{https://doi.org/10.1101/2020.04.17.20065441}
\showDOI{\tempurl}


\bibitem[\protect\citeauthoryear{Paszke, Gross, Massa, Lerer, Bradbury, Chanan,
  Killeen, Lin, Gimelshein, Antiga, Desmaison, Kopf, Yang, DeVito, Raison,
  Tejani, Chilamkurthy, Steiner, Fang, Bai, and Chintala}{Paszke
  et~al\mbox{.}}{2019}]%
        {pytorch}
\bibfield{author}{\bibinfo{person}{Adam Paszke}, \bibinfo{person}{Sam Gross},
  \bibinfo{person}{Francisco Massa}, \bibinfo{person}{Adam Lerer},
  \bibinfo{person}{James Bradbury}, \bibinfo{person}{Gregory Chanan},
  \bibinfo{person}{Trevor Killeen}, \bibinfo{person}{Zeming Lin},
  \bibinfo{person}{Natalia Gimelshein}, \bibinfo{person}{Luca Antiga},
  \bibinfo{person}{Alban Desmaison}, \bibinfo{person}{Andreas Kopf},
  \bibinfo{person}{Edward Yang}, \bibinfo{person}{Zachary DeVito},
  \bibinfo{person}{Martin Raison}, \bibinfo{person}{Alykhan Tejani},
  \bibinfo{person}{Sasank Chilamkurthy}, \bibinfo{person}{Benoit Steiner},
  \bibinfo{person}{Lu Fang}, \bibinfo{person}{Junjie Bai}, {and}
  \bibinfo{person}{Soumith Chintala}.} \bibinfo{year}{2019}\natexlab{}.
\newblock \showarticletitle{PyTorch: An Imperative Style, High-Performance Deep
  Learning Library}.
\newblock In \bibinfo{booktitle}{\emph{Advances in Neural Information
  Processing Systems 32}}, \bibfield{editor}{\bibinfo{person}{H.~Wallach},
  \bibinfo{person}{H.~Larochelle}, \bibinfo{person}{A.~Beygelzimer},
  \bibinfo{person}{F.~d\textquotesingle Alch\'{e}-Buc},
  \bibinfo{person}{E.~Fox}, {and} \bibinfo{person}{R.~Garnett}} (Eds.).
  \bibinfo{publisher}{Curran Associates, Inc.}, \bibinfo{pages}{8024--8035}.
\newblock
\urldef\tempurl%
\url{http://papers.neurips.cc/paper/9015-pytorch-an-imperative-style-high-performance-deep-learning-library.pdf}
\showURL{%
\tempurl}


\bibitem[\protect\citeauthoryear{ramesh babu, Zadeh, Jaiswal, Lueckenhoff,
  Kyrarini, and Makedon}{ramesh babu et~al\mbox{.}}{2020}]%
        {icmi20}
\bibfield{author}{\bibinfo{person}{Ashwin ramesh babu},
  \bibinfo{person}{Mohammad Zadeh}, \bibinfo{person}{Ashish Jaiswal},
  \bibinfo{person}{Alexis Lueckenhoff}, \bibinfo{person}{Maria Kyrarini}, {and}
  \bibinfo{person}{Fillia Makedon}.} \bibinfo{year}{2020}\natexlab{}.
\newblock \showarticletitle{A Multi-modal System to Assess Cognition in
  Children from their Physical Movements}.
\newblock
\urldef\tempurl%
\url{https://doi.org/10.1145/3382507.3418829}
\showDOI{\tempurl}


\bibitem[\protect\citeauthoryear{Shahroudy, Liu, Ng, and Wang}{Shahroudy
  et~al\mbox{.}}{2016}]%
        {ntu1}
\bibfield{author}{\bibinfo{person}{Amir Shahroudy}, \bibinfo{person}{Jun Liu},
  \bibinfo{person}{Tian-Tsong Ng}, {and} \bibinfo{person}{Gang Wang}.}
  \bibinfo{year}{2016}\natexlab{}.
\newblock \showarticletitle{NTU RGB+D: A Large Scale Dataset for 3D Human
  Activity Analysis}. In \bibinfo{booktitle}{\emph{IEEE Conference on Computer
  Vision and Pattern Recognition}}.
\newblock


\bibitem[\protect\citeauthoryear{Sánchez-DelaCruz, Weber, Biswal, Mejia,
  Hernández-Chan, and Gómez-Pozos}{Sánchez-DelaCruz et~al\mbox{.}}{2019}]%
        {gait19}
\bibfield{author}{\bibinfo{person}{Eddy Sánchez-DelaCruz},
  \bibinfo{person}{Roberto Weber}, \bibinfo{person}{Rajesh Biswal},
  \bibinfo{person}{Jose Mejia}, \bibinfo{person}{Gandhi Hernández-Chan}, {and}
  \bibinfo{person}{Heberto Gómez-Pozos}.} \bibinfo{year}{2019}\natexlab{}.
\newblock \showarticletitle{Gait Biomarkers Classification by Combining
  Assembled Algorithms and Deep Learning: Results of a Local Study}.
\newblock \bibinfo{journal}{\emph{Computational and Mathematical Methods in
  Medicine}}  \bibinfo{volume}{2019} (\bibinfo{date}{12} \bibinfo{year}{2019}),
  \bibinfo{pages}{1--14}.
\newblock
\urldef\tempurl%
\url{https://doi.org/10.1155/2019/3515268}
\showDOI{\tempurl}


\bibitem[\protect\citeauthoryear{Zadeh, Babu, Jaiswal, and Makedon}{Zadeh
  et~al\mbox{.}}{2020}]%
        {iui20}
\bibfield{author}{\bibinfo{person}{Mohammad~Zaki Zadeh},
  \bibinfo{person}{Ashwin~Ramesh Babu}, \bibinfo{person}{Ashish Jaiswal}, {and}
  \bibinfo{person}{Fillia Makedon}.} \bibinfo{year}{2020}\natexlab{}.
\newblock \bibinfo{title}{Self-Supervised Human Activity Recognition by
  Augmenting Generative Adversarial Networks}.
\newblock
\newblock
\showeprint[arxiv]{2008.11755}~[cs.CV]


\end{thebibliography}


\end{document}